\documentclass{llncs}

\usepackage{graphicx}
\usepackage{subfigure}
\usepackage{caption}
\usepackage{makeidx}
\usepackage{amsmath,amssymb,theorem,bm,exscale}
\usepackage{textcomp,bm}
\usepackage[latin1]{inputenc}
\usepackage{latexsym}
\usepackage{placeins}
\usepackage{color}
\usepackage{lineno}
\usepackage{array}
\usepackage{multirow}
\usepackage{enumitem}
\usepackage{dsfont}
\usepackage[pagebackref=false,breaklinks=true,letterpaper=true,colorlinks,urlcolor=blue,citecolor=blue,linkcolor=blue,bookmarks=false]{hyperref}

\begin{document}

\title{GREEN: a Graph REsidual rE-ranking Network \\for Grading Diabetic Retinopathy}
\titlerunning{Graph Residual Re-ranking Network}
\author{Shaoteng Liu\inst{1,2} \and Lijun Gong\inst{1}\thanks{L. Gong is the corresponding author. This work was done when S. Liu was an intern in Tencent Jarvis Lab.} \and
Kai Ma\inst{1} \and Yefeng Zheng\inst{1}}
\authorrunning{Liu et al.}
\institute{Tencent Jarvis Lab, Shenzhen, China\\
\and  Northwestern Polytechnical University, Xian, China\\
\email{lijungong@tencent.com}}

\maketitle              
\begin{abstract}
The automatic grading of diabetic retinopathy (DR) facilitates medical diagnosis for both patients and physicians.
Existing researches formulate DR grading as an image classification problem.
As the stages/categories of DR correlate with each other, the relationship between different classes cannot be explicitly described via a one-hot label because it is empirically estimated by different physicians with different outcomes.
This class correlation limits existing networks to achieve effective classification.
In this paper, we propose a Graph REsidual rE-ranking Network (GREEN) to introduce a class dependency prior into the original image classification network.
The class dependency prior is represented by a graph convolutional network with an adjacency matrix.
This prior augments image classification pipeline by re-ranking classification results in a residual aggregation manner.
Experiments on the standard benchmarks have shown that GREEN performs favorably against state-of-the-art approaches.

\end{abstract}
\section{Introduction}
Diabetic retinopathy (DR) is a common chronic disease leading to visual loss and blindness~\cite{cho2018idf}.
According to the severity of retinopathy lesion, DR is normally graded into five stages.\footnote{The five stages are defined as none, mild, moderate, severe and proliferative stages according to the International Clinical Diabetic Retinopathy scale~\cite{levels2002international}.}
Medical treatments for DR varies according to different DR grades~\cite{chakrabarti2012diabetic}.
In practice, DR grading is an empirical process executed by physicians, which requires sufficient expertise and time-consuming identifications.
Therefore, there is a need to develop automatic DR grading systems to benefit both patients and physicians for efficient diagnosis.

The convolutional neural networks (CNNs) have improved the grading performance of DR~\cite{gulshan2016development}.
Existing methods \cite{gulshan2016development,li2019canet,rakhlin2018diabetic} formulate the DR grading as an image classification task where each predefined category represents one stage of DR. Li et al.~\cite{li2019canet} proposed two attention modules (i.e., disease specific and disease dependent models) which were integrated into the CNN to automatically grade DR.
A number of randomly augmented images were generated in~\cite{rakhlin2018diabetic} for each sample and the CNN predictions of generated images were fused for final diagnosis. CNN predictions of fundus images captured from paired eyes (i.e., the corresponding left and right eyes) could be fused together for final predictions as well.
These DR grading systems typically use off-the-shelf CNNs (i.e., VGG~\cite{simonyan2014very} and ResNet~\cite{he2016deep}) which are designed for natural image classifications where the categories of natural images (i.e., ImageNet~\cite{deng2009imagenet}) usually do not correlate with each other.
Differently, fundus images of DR dataset are visually similar.
Besides, one stage/category of DR correlates with others, which brings difficulty for physicians to produce one-hot labels.\footnote{In each one-hot label, there is only one element whose value is 1 while the remaining elements are all 0.}
Fig.~\ref{fig:example} shows four examples of DR which were labeled differently by two physicians.
Moreover, a physician may label the same fundus image differently at different time.
When using the challenging training data to learn a classification network, we observe that the inter-class correlation is not effectively modeled and the classification performance is limited.

\renewcommand{\tabcolsep}{1pt}
\def\swone{0.2\linewidth}
\begin{figure}[t]
\centering
\begin{tabular}{cccc}
    \includegraphics[width=\swone]{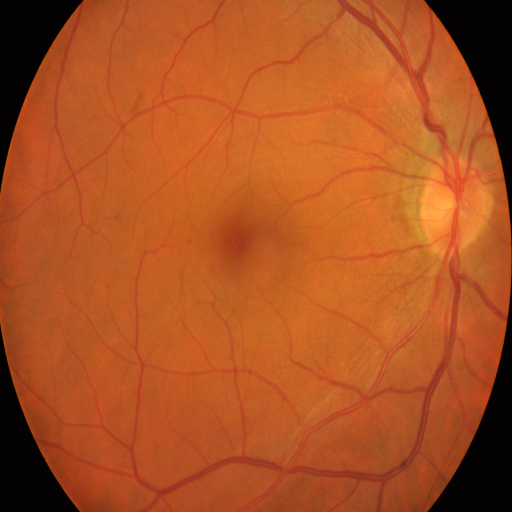}&
    \includegraphics[width=\swone]{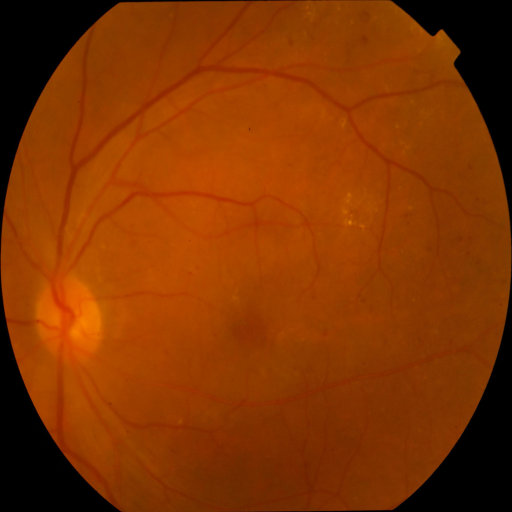}&
    \includegraphics[width=\swone]{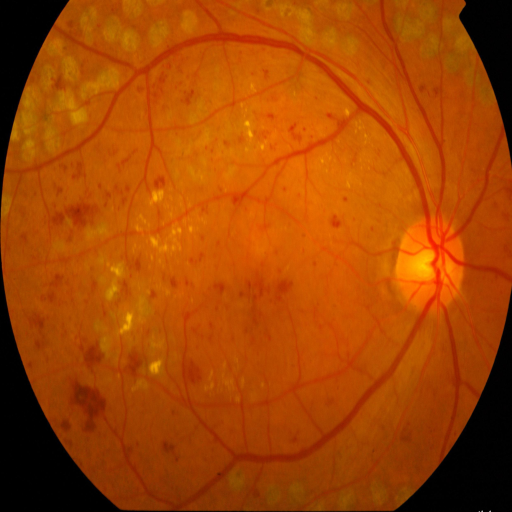}&
    \includegraphics[width=\swone]{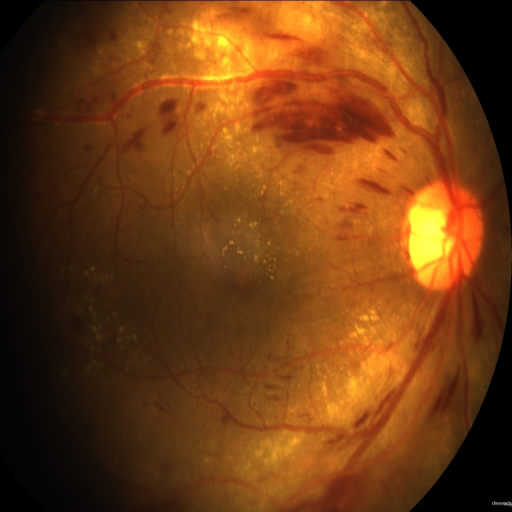}\\
    None (1)& Mild (1)& Severe (1)& Proliferative (1)\\
    Mild (2)& Moderate (2)& Moderate (2)& Severe (2)\\
\end{tabular}
\caption{Inconsistent labels from two physicians (i.e., 1 and 2) for the same DR fundus images. Class correlation causes this inconsistency and could limit the network classification accuracy.}
\label{fig:example}
\end{figure}

Soft labels are intuitively considered to model inter-class correlations. Bagherinezhad et al.~\cite{bagherinezhad2018label} proposed a label refinery approach to iteratively update ground truth labels during the training process. A label embedding network was designed in~\cite{sun2017label} to learn soft distributions of network predictions.
To the best of our knowledge, all existing methods tend to dynamically transform the one-hot label into a discrete probability distribution during the training process, and taking advantage of the inter-class correlation information, while leaving the structure of the classification network fixed.
However, the inter-class correlation is implicitly modeled within the classification network, which may hamper the network's convergence or bring limited improvement as shown in Table~\ref{Tab:ablation-DRD}.

In this paper, we propose a Graph REsidual rE-ranking Network (GREEN) to explicitly model the class correlation for significant DR grading improvement.
GREEN consists of a standard image classification network and an extra class-dependency module.
In this class-dependency module, we use a Graph Convolutional Network (GCN)~\cite{kipf2016semi} to integrate the class-dependency prior into the classification network.
The graph structure formulates each class as one node and models the relationship between different nodes via an adjacency matrix.
The adjacency matrix is learned to represent the dependencies of different classes in the CNN feature space.
The class-dependency module is trained end-to-end together with the image classification network.
The output of this module is kept fixed during inference and re-ranks classification results via residual aggregation.
To this end, the class-dependency is learned offline and functions as a constant prior when we classify DR images.
Experiments demonstrate that the proposed method improves original classification accuracy and performs favorably against state-of-the-art approaches.

\begin{figure}[t]
\centering
\includegraphics[width=0.75\textwidth]{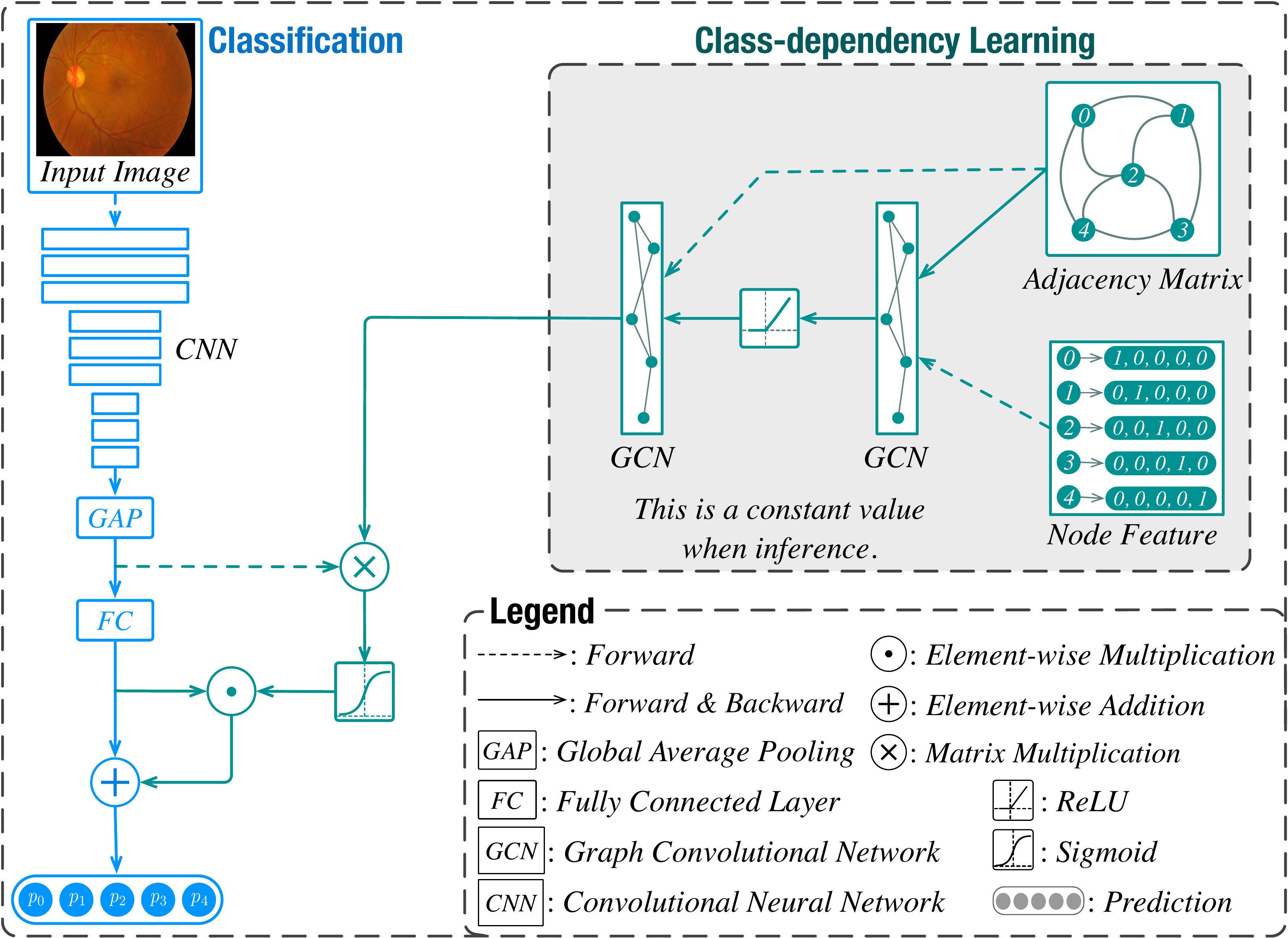}
\caption{Overview of the proposed framework. Besides the image classification network which consists of a feature backbone and a classifier, we propose a class-dependency module to model class correlation. The proposed module benefits image classification via residual aggregation to re-rank network prediction results.}
\label{fig:pipeline}
\end{figure}

\section{Proposed Method}
We illustrate the proposed Graph REsidual rE-ranking Network (GREEN) in Fig.~\ref{fig:pipeline} as an overview.
The input of GREEN is a fundus image and the output is the corresponding grade of DR.
The image classification network follows existing structures \cite{he2016deep,tan2019efficientnet} which includes a feature extraction backbone and a classifier.
On the other hand, the class-dependency module formulates class dependency as prior information to re-rank classification results via residual aggregation.
In the following, we first show how the class-dependency module improves network prediction via residual re-ranking. Then, we illustrate the details of the proposed module including architectures and training details.

\subsection{Image Classification with Residual Re-ranking}
The training data with label uncertainty diminishes CNNs' discriminative capability when grading DR. To mitigate this effect, we propose a class-dependency module to reweigh network predictions. The proposed module is learned offline and kept fixed during online inference. It provides constant values to fuse with input CNN features for classification re-ranking. We illustrate the original image classification network in the following and show how the proposed module re-ranks in details.

\subsubsection{Image Classification.}
A typical image classification network consists of a feature extraction backbone and a classifier. After extracting the CNN feature maps of an input fundus image, we apply a global average pooling (GAP) layer to the extracted feature maps and pass the output to the classifier (i.e., fully connected layers).
The advantage of using GAP is that it effectively maps an input vector with arbitrary dimension to a fixed one. As such, we do not need to resize input images to a fixed resolution (e.g., $256\times 256$), which is often a required preprocessing step of natural images. This benefits the severity estimation process which is akin to fine-grained image classification where the input fundus images are similar to each other. We keep the input images in high resolution to avoid potential detail missing to reduce misclassification.

\subsubsection{Residual Re-ranking.}
For an input fundus image $I$, the output of GAP is a vector with fixed dimension $d$. We denote this vector as $G \in \mathds{R}^{1\times d}$. The class-dependency module produces a learned matrix $C \in \mathds{R}^{n\times d}$ where $n$ is the number of output categories. We generate the instance re-ranking weight by fusing these two inputs. The fusion process is a matrix multiplication operation followed by a sigmoid activation, which can be written as:
\begin{equation}\label{eq:1}
  R_{\rm I}=\sigma \left(G\times C^{T}\right)
\end{equation}
where $C^{T}$ is the transpose matrix of $C$, $\sigma$ is the sigmoid function and $R_{\rm I}$ is the instance re-ranking weight. This operation projects $G$ into $C$ along the direction of each output category, and measures the corresponding correlation values. After computing $R_{\rm I}$, we re-rank the classification results via residual aggregation as follows:
\begin{equation}\label{eq:2}
  P_{\rm I}=\operatorname{Softmax} \left(R_{\rm I}\odot S_{\rm I}+S_{\rm I}\right)
\end{equation}
where $P_{\rm I}$ is the probability of output classes, $S_{\rm I}$ is the class prediction score from the fully connected (FC) layer and $\odot$ is the element-wise multiplication operation. Along with the original prediction $S_{\rm I}$ from the FC output, we set the re-rank weights on it to adjust the prediction scores. The $R_{\rm I}\odot S_{\rm I}$ term functions as an auxiliary prediction from the perspective of class-dependency prior on the current input.

\subsection{Graph Convolutional Network}

The proposed class-dependency module consists of a graph convolutional network (GCN). GCN is introduced in \cite{kipf2016semi} to perform semi-supervised classification, where the output of GCN is the probability of each class. Differently, we offline train the GCN to predict a constant class dependency prior to fuse with CNN features for final predictions.

The graph structure formulates each class as one node and constructs the relationship between different nodes via an adjacency matrix. As shown in Fig.~\ref{fig:pipeline}, the adjacency matrix, together with node feature, is the input of GCN. The output is a learned projection matrix that contains the representation of each class in the CNN feature space. The dimension of this feature space is same as that of GAP output. For each input image, the corresponding feature representation after GAP output is multiplied with the projection matrix. The multiplication results indicate the correlation values between input features and each class.

Following~\cite{kipf2016semi}, the GCN adopted in the proposed module consists of two matrix multiplication layers with a ReLU activation function. We denote the adjacency matrix as $A$, node features as $X$, and the weight of the first layer as $W_1$. The output of the first layer can be written as:
\begin{equation}\label{eq:3}
  \mathcal{C}_1(A,X,W_1)=A\times X\times W_1
\end{equation}
where each row of $A\in \mathds{R}^{n\times n}$ is initialized following the Gaussian distribution with the peak set on the diagonal position, $X\in \mathds{R}^{n\times n}$ is an identical matrix and represents there are $n$ nodes in the graph. $W_1\in \mathds{R}^{n\times h}$ maps dimension from $n$ to $h$ (the hidden layer). We send $\mathcal{C}_1(A,X,W_1)$ to the remaining GCN structures and obtain the output as:
\begin{equation}\label{eq:4}
  C=A\times \operatorname{ReLU}\left(\mathcal{C}_1\right)\times W_2
\end{equation}
where $C\in \mathds{R}^{n\times d}$ is the prior knowledge of the class dependency and $W_2\in \mathds{R}^{h\times d}$ maps dimension from the hidden layer to output class. Thus, we can learn the complex inter-class correlations by stacking two GCN layers. We refer interested readers to~\cite{kipf2016semi} for more details.
During the training stage, we fuse $C$ with $G$ via Eq.~\eqref{eq:1} for forward computation. The gradients via back propagation are passed into both GCN and image classification network. When learning GCN, we update both adjacent matrix $A$ and GCN parameters (i.e., $W_1$ and $W_2$) as they constitute the class-dependency prior. This class-dependency module is trained end-to-end together with the image classification network and the output is kept fixed as a constant prior during inference.

\begin{figure}[t]
\centering
\includegraphics[width=0.69\textwidth]{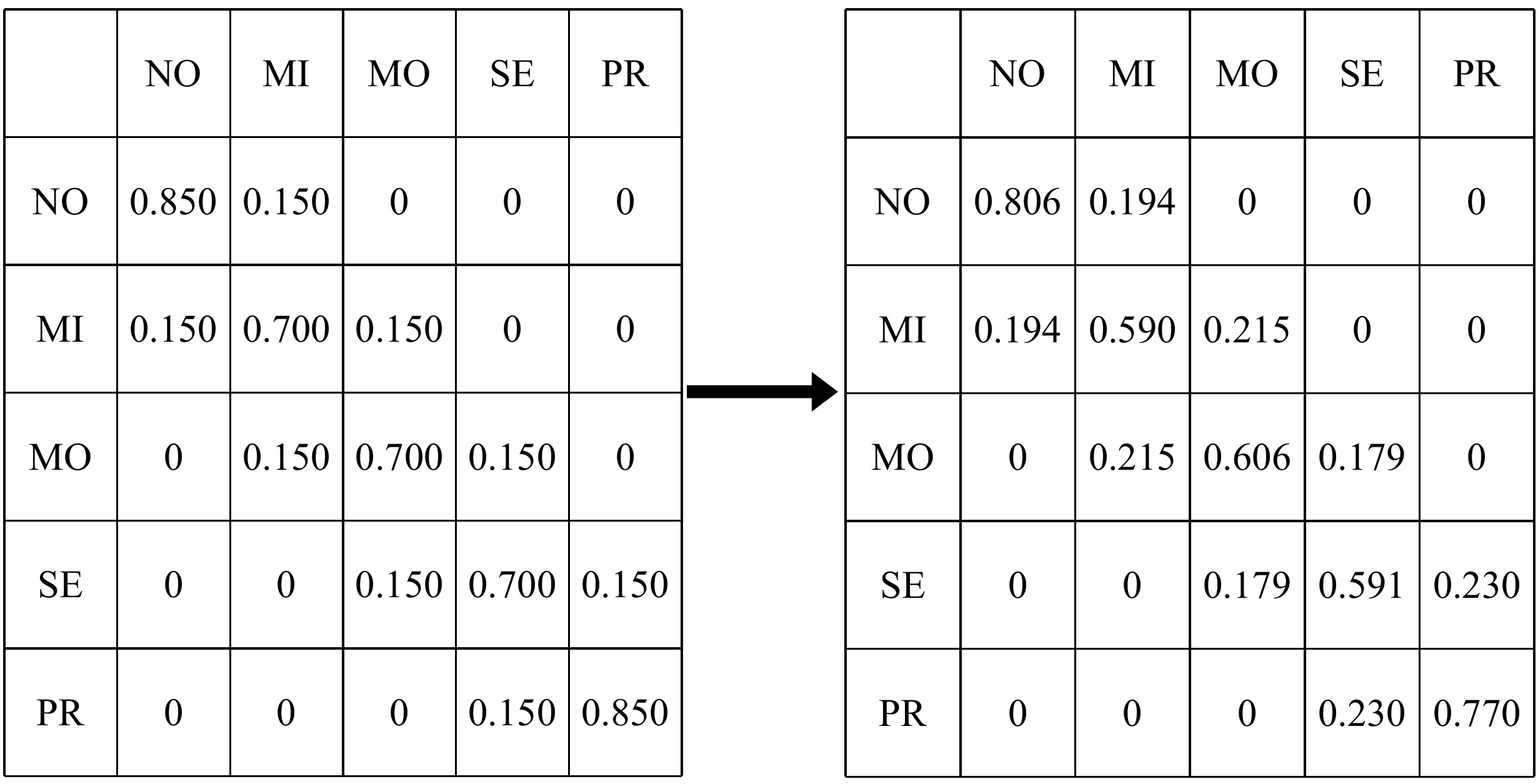}
\caption{Variations of elements in adjacency matrix before and after training. The abbreviation NO, MI, MO, SE and PR indicate corresponding five stages of DR. The learned matrix is fixed during testing to consistently provide class dependency priors for DR grading.}
\label{fig:matrix}
\end{figure}

\subsubsection{Adjacency matrix visualization.}

Fig.~\ref{fig:matrix} shows the adjacency matrix representing class-dependency. The matrix shown on the left is the initial matrix where the abbreviation NO, MI, MO, SE and PR correspond to none, mild, moderate, severe and proliferative stages respectively.
The elements along the diagonal are initialized with large values while their neighbors are set to relatively small values.
The values of neighboring elements indicate the class dependency between different categories. After training, we observe that in the matrix shown on the right part of Fig.~\ref{fig:matrix}, the values with distant categories are 0 (e.g., the adjacent value between NO and MO, or MI and PR). This shows that there exists a clear boundary between distant categories. Furthermore, the adjacent values between MI and MO, SE and PR are higher than others.
It complies with the empirical observation from clinicians that the boundary of mild and moderate stages is not obvious, so as the severe and proliferative stages.

\section{Experiments}

In this section, we evaluate the proposed method on two DR benchmark datasets: Diabetic Retinopathy Detection (DRD)~\cite{DRD} and APTOS 2019 Blindness Detection (APTOS2019)~\cite{APTOS2019}. There are 35,126 and 3,662 fundus images in DRD and APTOS2019, respectively. We split each dataset into five folds for cross validation and use weighted kappa, weighted accuracy, and weighted F1 score as evaluation metrics.

We use EfficientNet-b0~\cite{tan2019efficientnet} as image classification baseline with ImageNet pretraining~\cite{deng2009imagenet}. When training with fundus images, we utilize SGD as the optimizer and set the initial learning rate as $1e^{-3}$. The batch size is set as 128 and the training procedure ends at the $60^{th}$ epoch. More results are provided in the supplementary file. We will make our implementation code available to the public.

\subsection{Ablation Study}
The contribution of GREEN is to use a class-dependency module for classification reweighing. In this ablation study, we show how this module improves baseline classification accuracies. Meanwhile, training with soft labels (i.e., LEN~\cite{sun2017label} and LRN~\cite{bagherinezhad2018label}) and training with fixed adjacency matrix~\cite{kipf2016semi} are also employed with the baseline network for comparisons. The evaluations are conducted on both datasets.

Table~\ref{Tab:ablation-DRD} shows the experimental results on the DRD dataset with EfficientNet-b0~\cite{tan2019efficientnet} as the backbone for experimental efficiency. Compared to the baseline performance, the soft label learning schemes (i.e., LEN and LRN) bring limited improvement under all metrics. On the other hand, training GREEN with a fixed adjacency matrix (as shown in the left part of Fig.~\ref{fig:matrix}) does not robustly outperform soft label learning schemes (e.g., 0.773 v.s. 0.776 of F1 score). Nevertheless, by training the proposed class-dependency module with the adjacency matrix end-to-end, we consistently improve the baseline and perform favorably against other learning configurations under all metrics.
Similar performance on APTOS2019 dataset is shown in supplementary file where GREEN performs favorably against other configurations on baseline improvements.

\renewcommand{\tabcolsep}{3pt}
\begin{table}[t]
\centering
\caption{The ablation study with different training schemes on the DRD dataset. GREEN performs favorably against other methods on baseline improvements.}
\label{Tab:ablation-DRD}
\small
\begin{tabular}{m{3.5cm}m{2.5cm}m{1.5cm}<{\centering}m{1.5cm}<{\centering}m{1.5cm}<{\centering}}
\hline
\multirow{2}{*}{Method} & \multirow{2}{*}{Backbone} & \multicolumn{3}{c}{Evaluation Metrics}\\
\cline{3-5}
 &  & Kappa & Accuracy & F1 score \\
\hline
Base & EfficientNet-b0 & $0.583$ & $0.803$	& $0.754$ \\
Base + LRN~\cite{bagherinezhad2018label} & EfficientNet-b0 & $0.619$ & $ 0.810$ & $0.776$ \\
Base + LEN~\cite{sun2017label} & EfficientNet-b0 & $0.623$ & $ 0.808$ & $0.764$ \\
Base + GREEN (fixed adjacency matrix) & EfficientNet-b0 & $0.683$ & $0.811$ & $0.773$ \\
Base + GREEN & EfficientNet-b0 & $\textbf{0.700}$ & $\textbf{0.816}$ & $\textbf{0.782}$ \\
\hline
\end{tabular}
\end{table}

\subsection{Comparisons with state-of-the-art approaches}
We also compare GREEN with state-of-the-art approaches including DLI~\cite{rakhlin2018diabetic} and CANet~\cite{li2019canet}. DLI fuses multiple sources (i.e., randomly augmentation images and paired eyes if possible) while CANet involves deep CNN attentions. For a fair comparison, we follow CANet to use ResNet-50 as the feature extraction backbone. Besides, we also validate the effectiveness of GREEN by using prevalent backbones (i.e., EfficientNet-b0 and Se-ResNeXt50).

Table~\ref{Tab:comparison-DRD} shows the results on the DRD dataset. GREEN consistently outperforms other methods under all the evaluation metrics by using ResNet-50. Furthermore, the results show the effectiveness of GREEN of EfficientNet-b0 and Se-ResNeXt50 backbones. In Table~\ref{Tab:comparison-APTOS}, the evaluation results are similar on the APTOS2019 dataset. With the same CNN backbone, GREEN performs favorably against DLI and CANet. By using other CNN backbones, GREEN is shown to be effective as well. The evaluations on these two datasets indicate that GREEN suits prevalent backbones for effective DR grading and performs favorably against existing methods.

\renewcommand{\tabcolsep}{1pt}
\def\swone{0.22\linewidth}
\begin{figure}[t]
\scriptsize
\centering
\begin{tabular}{cccc}
    \includegraphics[width=\swone]{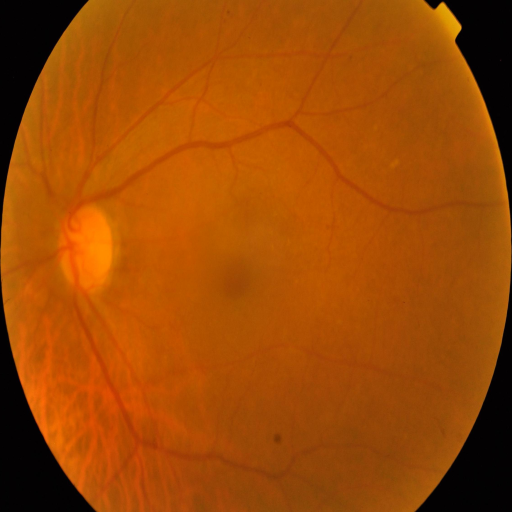}&
    \includegraphics[width=\swone]{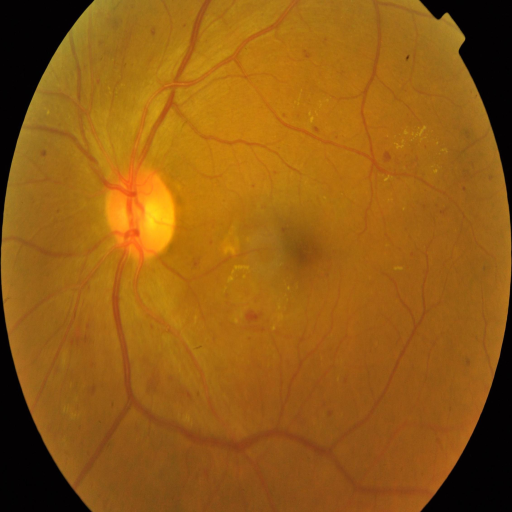}&
    \includegraphics[width=\swone]{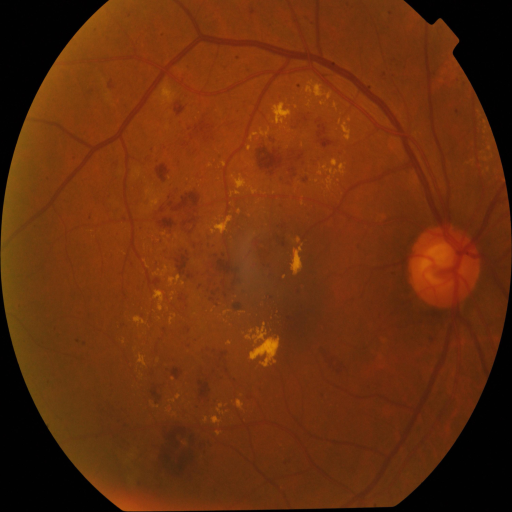}&
    \includegraphics[width=\swone]{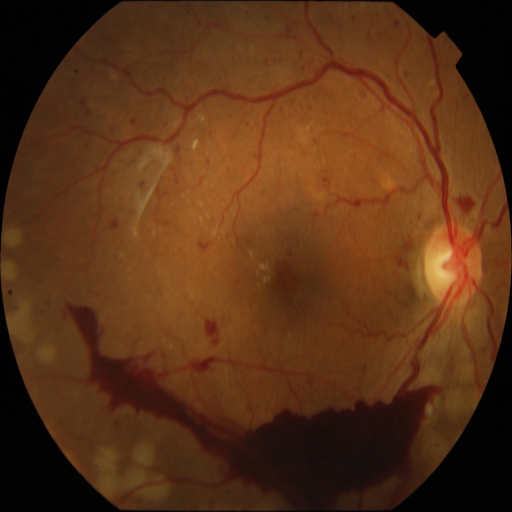}\\
    DIL / CANet: NO / NO& DIL / CANet: MI / MO& DIL / CANet: MO / SE& DIL / CANet: PR / PR\\
    GREEN / GT: MI / MI & GREEN / GT: MO / MO & GREEN / GT: MO / MO& GREEN / GT: SE / SE\\
\end{tabular}
\caption{Visual evaluations on the APTOS2019 dataset. GREEN is compared to DIL, CANet, and the ground truth (GT). The abbreviation NO, MI, MO, SE, PR correspond to the five DR stages illustrated in~\cite{levels2002international}. The results show GREEN is effective to perform DR grading. }
\label{fig:ex}
\end{figure}

\begin{table}[t]
\centering
\caption{Comparison with state-of-the-art on the DRD~\cite{DRD} dataset.}
\label{Tab:comparison-DRD}
\small
\begin{tabular}{m{2.5cm}m{2.5cm}m{1.5cm}<{\centering}m{1.5cm}<{\centering}m{1.5cm}<{\centering}}
\hline
\multirow{2}{*}{Method} & \multirow{2}{*}{Backbone} & \multicolumn{3}{c}{Evaluation Metrics}\\
\cline{3-5}
 &  & Kappa & Accuracy & F1 score \\
\hline
DLI~\cite{rakhlin2018diabetic} & ResNet50 & $0.620$ & $0.809$	& $0.765$ \\
CANet~\cite{li2019canet} & ResNet50 & $0.649$ & $0.816$ & $0.774$ \\
GREEN & ResNet50 & $0.693$ & $0.820$ & $0.780$ \\
GREEN & EfficientNet-b0 & 0.700 & 0.816 & 0.782 \\
GREEN & Se-ResNeXt50 & $\textbf{0.727}$ & $\textbf{0.826}$ & $\textbf{0.790}$ \\
\hline
\end{tabular}
\centering
\caption{Comparison with state-of-the-art on the APTOS2019~\cite{APTOS2019} dataset.}
\label{Tab:comparison-APTOS}
\small
\begin{tabular}{m{2.5cm}m{2.5cm}m{1.5cm}<{\centering}m{1.5cm}<{\centering}m{1.5cm}<{\centering}}
\hline
\multirow{2}{*}{Method} & \multirow{2}{*}{Backbone} & \multicolumn{3}{c}{Evaluation Metrics}\\
\cline{3-5}
 &  & Kappa & Accuracy & F1 score \\
\hline
DLI~\cite{rakhlin2018diabetic} & ResNet50 & $0.895$ & $0.825$	& $0.803$ \\
CANet~\cite{li2019canet} & ResNet50 & $0.900$ & $0.832$ & $0.813$ \\
GREEN & ResNet50 & $0.908$ & $0.844$ & $0.836$ \\
GREEN & EfficientNet-b0 & 0.910 & 0.848 & 0.835 \\
GREEN & Se-ResNeXt50 & $\textbf{0.912}$ & $\textbf{0.857}$ & $\textbf{0.852}$ \\
\hline
\end{tabular}
\end{table}

Fig.~\ref{fig:ex} shows the visual evaluation results. We compare GREEN with DIL and CANet on the APTOS2019 dataset. In general, DIL and CANet is not effective to accurately grade DR over all stages. This is due to a lack of class-dependency modeling during CNN classification. In comparison, by formulating class-dependency with GCN and integrating it into the CNN for end-to-end training, we achieve favorable results when grading DR of all the stages.

\section{Concluding Remarks}

In this work, we proposed to model class correlations for DR grading via a class-dependency formulation. In the proposed model, we integrated GCN into the original classification network and trained the whole network end-to-end. This model was kept fixed during online inference to consistently produce class dependency priors, which re-ranked the original classification results via residual aggregation. Experiments on the benchmark datasets showed the effectiveness of the proposed method by using prevalent CNN backbones. Meanwhile, the proposed method performed favorably against state-of-the-art approaches.
\\
\\
\textbf{Acknowledgments.} This work was funded by the Key Area Research and Development Program of Guangdong Province, China (No. 2018B010111001), National Key Research and Development Project (2018YFC2000702) and Science and Technology Program of Shenzhen, China (No. ZDSYS201802021814180).
%
%
\bibliographystyle{splncs04}
\bibliography{ref}

\begin{thebibliography}{10}
\providecommand{\url}[1]{\texttt{#1}}
\providecommand{\urlprefix}{URL }
\providecommand{\doi}[1]{https://doi.org/#1}

\bibitem{DRD}
Diabetic Retinopathy Detection (2016),
  \url{https://www.kaggle.com/c/diabetic-retinopathy-detection/data}

\bibitem{APTOS2019}
APTOS 2019 Blindness Detection (2019),
  \url{https://www.kaggle.com/c/aptos2019-blindness-detection/data}

\bibitem{levels2002international}
Association, E.M., et~al.: International clinical diabetic retinopathy disease
  severity scale, detailed table  (2002),
  \url{http://www.icoph.org/dynamic/attachments/resources/diabetic-retinopathy-detail.pdf}

\bibitem{bagherinezhad2018label}
Bagherinezhad, H., Horton, M., Rastegari, M., Farhadi, A.: Label refinery:
  Improving imagenet classification through label progression. arXiv preprint
  arXiv:1805.02641  (2018)

\bibitem{chakrabarti2012diabetic}
Chakrabarti, R., Harper, C.A., Keeffe, J.E.: Diabetic retinopathy management
  guidelines. Expert Review of Ophthalmology  \textbf{7}(5),  417--439 (2012)

\bibitem{cho2018idf}
Cho, N., Shaw, J., Karuranga, S., Huang, Y., da~Rocha~Fernandes, J., Ohlrogge,
  A., Malanda, B.: Idf diabetes atlas: Global estimates of diabetes prevalence
  for 2017 and projections for 2045. Diabetes research and clinical practice
  \textbf{138},  271--281 (2018)

\bibitem{deng2009imagenet}
Deng, J., Dong, W., Socher, R., Li, L.J., Li, K., Fei-Fei, L.: Imagenet: A
  large-scale hierarchical image database. In: IEEE Conference on Computer
  Vision and Pattern Recognition. pp. 248--255. Ieee (2009)

\bibitem{gulshan2016development}
Gulshan, V., Peng, L., Coram, M., Stumpe, M.C., Wu, D., Narayanaswamy, A.,
  Venugopalan, S., Widner, K., Madams, T., Cuadros, J., et~al.: Development and
  validation of a deep learning algorithm for detection of diabetic retinopathy
  in retinal fundus photographs. Journal of the American Medical Association
  \textbf{316}(22),  2402--2410 (2016)

\bibitem{he2016deep}
He, K., Zhang, X., Ren, S., Sun, J.: Deep residual learning for image
  recognition. In: IEEE Conference on Computer Vision and Pattern Recognition.
  pp. 770--778 (2016)

\bibitem{kipf2016semi}
Kipf, T.N., Welling, M.: Semi-supervised classification with graph
  convolutional networks. arXiv preprint arXiv:1609.02907  (2016)

\bibitem{li2019canet}
Li, X., Hu, X., Yu, L., Zhu, L., Fu, C.W., Heng, P.A.: Canet: Cross-disease
  attention network for joint diabetic retinopathy and diabetic macular edema
  grading. IEEE Transactions on Medical Imaging  (2019)

\bibitem{rakhlin2018diabetic}
Rakhlin, A.: Diabetic retinopathy detection through integration of deep
  learning classification framework. bioRxiv p. 225508 (2018)

\bibitem{simonyan2014very}
Simonyan, K., Zisserman, A.: Very deep convolutional networks for large-scale
  image recognition. arXiv preprint arXiv:1409.1556  (2014)

\bibitem{sun2017label}
Sun, X., Wei, B., Ren, X., Ma, S.: Label embedding network: Learning label
  representation for soft training of deep networks. arXiv preprint
  arXiv:1710.10393  (2017)

\bibitem{tan2019efficientnet}
Tan, M., Le, Q.V.: Efficientnet: Rethinking model scaling for convolutional
  neural networks. arXiv preprint arXiv:1905.11946  (2019)

\end{thebibliography}

\end{document}